\documentclass[lettersize,journal]{IEEEtran}
\usepackage{amsmath,amsfonts}
\usepackage{algorithm}
\usepackage{array}
\usepackage[caption=false,font=normalsize,labelfont=sf,textfont=sf]{subfig}
\usepackage{textcomp}
\usepackage{stfloats}
\usepackage{url}
\usepackage{verbatim}
\usepackage{graphicx}
\usepackage{cite}
\usepackage[table,xcdraw]{xcolor}

\usepackage{algpseudocode}
\usepackage{multirow}
\usepackage{cite}
\usepackage{amsmath,amssymb,amsfonts}
\usepackage{graphicx}
\usepackage{textcomp}
\usepackage{xcolor}
\usepackage{color}
\usepackage{algorithm}
\usepackage{algpseudocode}
\usepackage{authblk}
\usepackage{hyperref}

\definecolor{blue}{rgb}{0, 0, 0}
\definecolor{blue1}{rgb}{0, 0, 0}

\newcommand{\tuba}[1]{\textcolor{blue}{#1}}
\newcommand{\tba}[1]{\textcolor{blue}{#1}}

\newcommand{\reviewtwo}[1]{\textcolor{blue1}{#1}}

\title{Multi-Object Graph Affordance Network: Goal-Oriented Planning through Learned Compound Object Affordances }

\author{Tuba Girgin$^{1,2}$ and Emre Uğur$^{1}$ \thanks{This work was supported by INVERSE project (101136067) funded by the European Union.} \thanks{$^1$Department of Computer Engineering, Bogazici University, Istanbul, Turkey. $^2$Robotics and Autonomous Systems Laboratory, TUBITAK BILGEM, 41470 Kocaeli, Turkey} }

\date{September 2023}

\begin{document}

\maketitle

\begin{abstract}
     Learning object affordances is an effective tool in the field of robot learning. While the data-driven models investigate affordances of single or paired objects, there is a gap in the \textit{exploration} of affordances of compound objects composed of an arbitrary number of objects. We propose the Multi-Object Graph Affordance Network which models complex compound object affordances by learning the outcomes of robot actions that facilitate interactions between an object and a compound. Given the depth images of the objects, the object features are extracted via convolution operations and encoded in the nodes of graph neural networks. Graph convolution operations are used to encode the state of the compounds, which are used as input to decoders to predict the outcome of the object-compound interactions. After learning the compound object affordances, given different tasks, the learned outcome predictors are used to plan sequences of stack actions that involve stacking objects on top of each other, inserting smaller objects into larger containers and passing through ring-like objects through poles. We showed that our system successfully modeled the affordances of compound objects that include concave and convex objects, in both simulated and real-world environments. We benchmarked our system with a baseline model to highlight its advantages. 
\end{abstract}

\begin{IEEEkeywords}
Robot learning, affordance learning, graph neural network (GNN) architecture.
\end{IEEEkeywords}

\section{Introduction}

\IEEEPARstart{T}{he} affordances concept, introduced by J.J. Gibson to refer to the action possibilities provided by the environment \cite{gibson1977theory}, has been significantly influential in robotics research \cite{rome2008macs,jamone2016affordances}. The developmental aspects of affordances have been widely adopted in robot learning research \cite{csahin2007afford, ugur2011goal, ugur2015staged,ugur2014emergent}. While previous works have examined the affordances of single or paired simple object interactions, the affordances of compound objects composed of an arbitrary number of objects with concave shapes and varying sizes have not been sufficiently studied \cite{Zech2017}.

Consider an infant trying to build a tower with its toys. Because of the different shapes and sizes of the objects, each toy would afford different actions, and with each action, various effects would be generated. The affordances of the objects may change according to their relations with the other objects in the environment, i.e., while an empty cup affords insertability, it might loose this affordance if one or more objects are put inside the cup or a large box is stacked on the cup. 
However, if a small object is inserted in a big cup or several large rings are stacked on the cup, the cup would remain insertable. Predicting the affordance of the compound object is not straightforward, as the affordance of a compound object not only depends on the affordances of the included objects but is also determined based on in which order these objects are placed (e.g. via releasing the objects on the compound) and the relative positions of all objects. In order to address the challenge of learning compound object affordances, we propose to represent the objects in the compound as a graph, as the graph representation preserves spatial relations between objects, and can be used to propagate information along the chain of objects in the compound, enabling effective reasoning for the complete structure. 

\begin{figure}
    \centering
    \includegraphics[ width=0.45\textwidth]{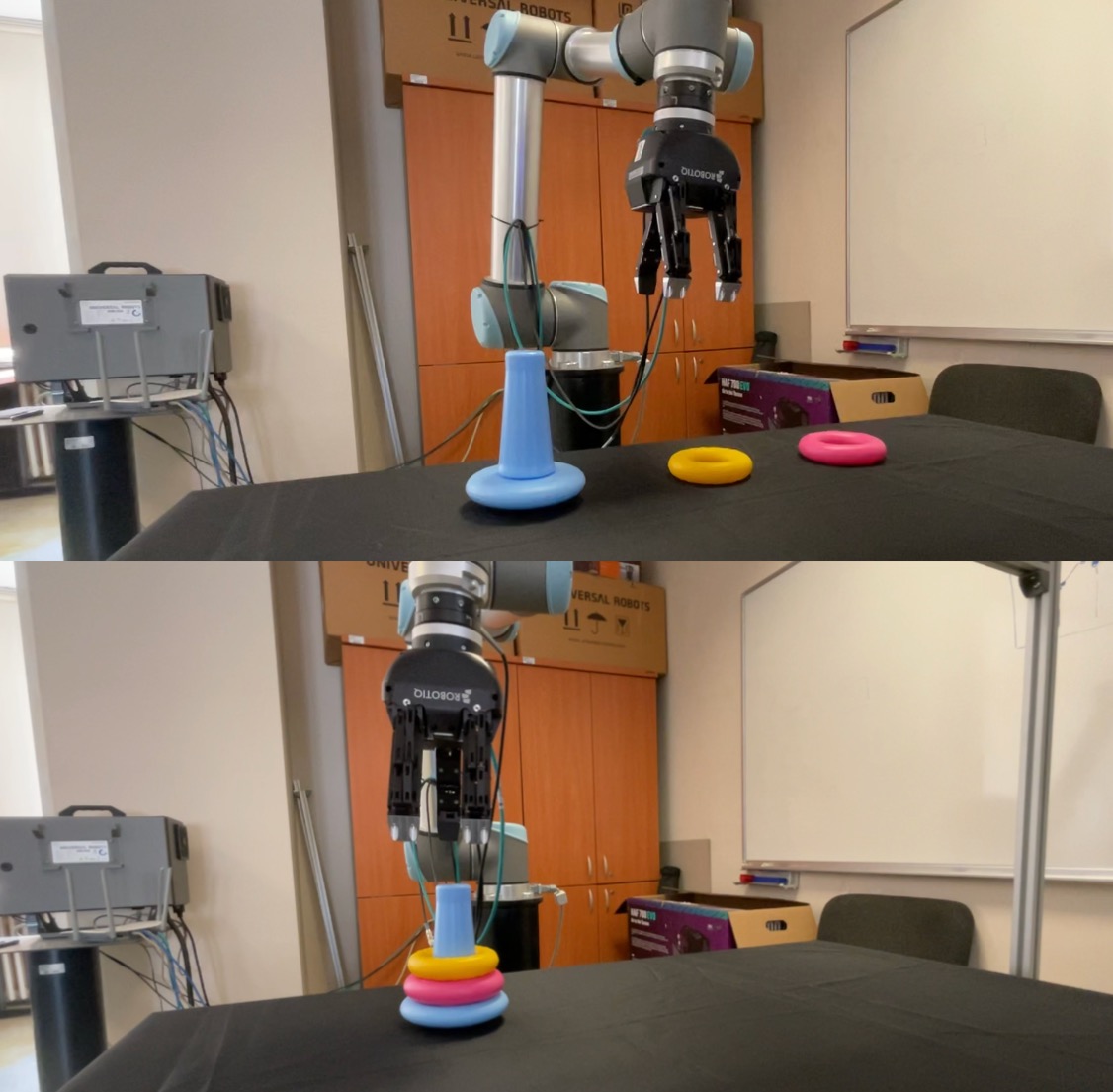}
    \caption{Execution of the plan generated using our MOGAN model to build the shortest compound object given a pole and two rings in the real world setup. \tuba{The agent uses the pole as the base and stacks the rings, as they do not change the height of the compound.}}
    \label{fig:real_ur10}
\end{figure}

Graph Neural Networks (GNNs) \cite{wu2020comprehensive} are effective for learning meaningful representations of structures and their relations. Consequently, they have gained extensive adoption in action recognition problems \cite{ahmad2021graph}, natural language processing \cite{wu2023graph}, navigation problems to learn relations between pedestrians, objects, and robots \cite{10115223, zhou2024learning, escudie2024}, as well as reasoning about relations between multi-object systems \cite{battaglia2016interaction, chang2016compositional, li2018propagation, tekden2021object}. In these studies, the representation capacity of GNNs for spatial relations involving an unlimited number of inputs is exploited, whereas other commonly used feed-forward networks lack this property. 

Affordances refer to the relations between objects, actions and effects \cite{csahin2007afford}, we aim to learn to predict effects given objects and actions. In our study, we propose the Multi-Object Graph Affordance Network (MOGAN), which learns affordances of compound objects, i.e. learns predicting effects of actions applied on objects and/or object compounds. The prediction is done using features obtained from graph representations utilizing GNNs. In this study, we focus on actions that facilitate interactions between objects. For this purpose, we used actions that pick up objects and place them on top of other objects or object compounds. The effects are encoded as the spatial displacements between the released objects and the objects in the compound structure. \tuba{With concave objects, visibility can be altered when stacking, as the line of sight may intersect another object. In contrast, with convex objects, there is no need to measure visibility. Consequently, when dealing with concave objects, traditional effect representation based solely on their center points is insufficient to reason about visibility effects. Therefore, more complex effect representations are required to accurately convey information about the visibility of concave objects. As a result, a suitable novel effect representation is used.} 



We designed six different tasks using an inventory of convex and concave objects of varying sizes, including poles, cups and rings of different sizes, boxes, and balls. The learned affordances correspond to forward predictors, and therefore can be used for goal-oriented action selection and  planning. We first presented the effect prediction results of compound object interactions in the Pybullet simulation environment. Subsequently, we discussed the success rates of plans generated through model predictions. Finally, we demonstrated the realization of the generated plans in the simulation environment, resulting in the construction of unseen structures composed of available objects. The results of our model were compared with those of the baseline model which corresponds to DeepSym model \cite{ahmetoglu2022deepsym}. While Deepsym is the state-of-the-art model for learning action-object-effect relational categories, it is limited to paired object interactions. We modified the Deepsym model to handle multi-object representations to benchmark it against our proposed model. We also demonstrated the applicability of our system by executing tasks with the UR10 manipulator in a real-world setting.

In summary, this paper introduces the MOGAN model, a novel approach for learning compound object affordances. Our contributions can be summarized as follows:

\begin{itemize}
  \item \textbf{Proposal of  Multi-Object Graph Affordance Network:} We introduce a novel model, MOGAN, which \textit{encodes}  the affordances of compounds composed of varying number of objects by representing them as a graph structure without the need for supervision from experts or labeled affordances. Our model learns these affordances through the observed effects of robot interactions.
  \item \textbf{Introduction of a Novel Effect Encoding Method:} We represent the effects of robot interactions with a particular encoding that takes into account 3D spatial relations. Widely used effects, such as displacement of the centers of the objects, are insufficient to explain the semantic behavior of concave objects.  
  \item \textbf{Demonstration of Applicability:}  We showed the applicability of our system by successfully accomplishing various tasks in both the Pybullet simulation environment and the real world using the UR10 manipulator.
\end{itemize}

\section{Related Work}

\subsection{Affordances}

 The study of affordances \cite{csahin2007afford,jamone2016affordances}, \tuba{\cite{yang2023recent,min2016affordance}} has attracted significant attention in recent years, with embodied AI studies utilizing affordances to evaluate language model generations. For instance, Ahn et al. \cite{ahn2022can} proposed SayCan, where they combined the skill affordances of a robot with Language Models (LLM) to ground the instructions to the environment. Additionally, Ahn et al. \cite{ahn2024autort} introduced AutoRT, where affordance filtering is utilized to align the tasks generated by vision-language models (VLMs) with the robot's capabilities and safety regulations. However, these studies primarily focused on extracting skill affordances by combining language model outputs with predefined instructions and rules, such as ``do not lift heavy objects." In contrast, the concept in our study revolves around exploring the affordances of objects through robot interactions and observed effects, which are influenced by also the weights of the objects.

Some approaches learned visual affordances to understand applicable actions through neural networks employed in computer vision. \cite{hassanin2021visual} Qian et al. \cite{qian2024affordancellm} combined large-scale vision language models with an image encoder and an affordance decoder network to predict an affordance map based on the queried action. Birr et al. \cite{birr2024autogpt+} extracted affordances of detected objects through queries using a predefined prompt list of affordances in ChatGPT, an AI tool developed by OpenAI.
Do et al. \cite{do2018affordancenet} detected affordances of objects in images along with their classes, with affordances being labeled at the pixel level. \tuba{Cuttano et al. \cite{cuttano2024does} proposed a model based on CLIP \cite{radford2021learning} that grounds task-agnostic affordances of texts from an open vocabulary onto images. Depth values of object images are also utilized to define affordance relations. Toumpa and Cohn \cite{toumpa2023object} defined affordance relations concerning the concaveness of objects. While they measure concaveness by inspecting depth values based on \cite{leyton1988process}, we use deep autoencoders to learn features of objects that are not limited to concavity.}

Various robotics research benefited affordances to enhance precision in grasping, picking, and placing operations \cite{ugur2014bootstrapping,hangl2016robotic,yang2021learning}.  Hart et al. \cite{hart2015affordance} designed a ROS package enabling the operators to specify grasp poses. Corona et al. \cite{corona2020ganhand} proposed a comprehensive network called GanHand, where the hand shape and pose are predicted through the reconstruction of the object alongside the predicted grasp type. \tuba{\cite{ugur2016emergent} studies mechanisms that produce hierarchical structuring of affordance learning tasks of different levels of complexity.}
 Mandikal and Grauman \cite{mandikal2021learning} learned grasping policies utilizing 3D thermal affordance maps. Zeng et al. \cite{zeng2022robotic} benefitted from labeled affordance maps for improved grasping. Learning contact information as affordances, as studied in \cite{schiavi2023learning} and \cite{geng2023rlafford}, is another way to tackle the existing problems.  Cheng et al. \cite{DBLP:conf/corl/ChengMS21} also learned the contact points of two objects for picking, grasping, and regrasping operations. Lin et al. \cite{lin2023mira} learned pixel-wise pick-and-place affordances by generating 3D imaginary scenes from 2D images using an annotated dataset.  Borja-Diaz et al. \cite{borja2022affordance} designed a self-supervised affordance learning model that labels gripper open and close points while the robot is controlled through human teleoperation. Mees et al. \cite{mees2023grounding} extended this work, grounding large language models to robotic applications. While these studies supervise their models to learn affordances using expert annotations, contact points, and gripper signals, we observe the effects of the manipulator's actions to explore the affordances. \tuba{ Cruz et al. \cite{cruz2016training} used affordances in an interactive reinforcement learning setup in order to speed up the learning of skills from other agents.}

Multiple approaches have studied the exploration of affordances learning effects through interactions \cite{moldovan2012learning, moldovan2014occluded}. Mar et al. \cite{mar2015self} explored affordance categories according to the effects of tool usage. They mapped the features extracted from the observations and affordance classes discovered by clustering the effects with the k-means algorithm \cite{macqueen1967some}. Antunes et al. \cite{antunes2016human} defined affordances as the probability of effects given the object features, tool features, and action. With the formulation of goals as symbols, they achieved probabilistic planning.  \tuba{Saponaro et al. \cite{saponaro2019beyond} exploit the affordances learned from robot interactions to interpret and describe human actions in light of their own experience.}
\cite{ugur2011unsupervised,ahmetoglu2022deepsym} also performed sub-symbolic and symbolic planning \cite{taniguchi2018symbol}, using affordances of only single or paired objects.

While the affordances of objects with concave and convex shapes, such as mugs and spoons, are learned with the supervision of experts, \textit{ exploring } and \textit{discovering} the affordances of complex structures generated by combining (inserting, passing through, stacking) of a sequence of such concave and convex objects through observed effects has not been studied to the best of our knowledge. We study the discovery of affordances of compounds, including concave shapes like rings, poles, and cups. In our study, we also adapt GNNs to exploit their representation capacity for compound object affordances.

\begin{figure*}[]
    \centering
    \includegraphics[width=17cm]{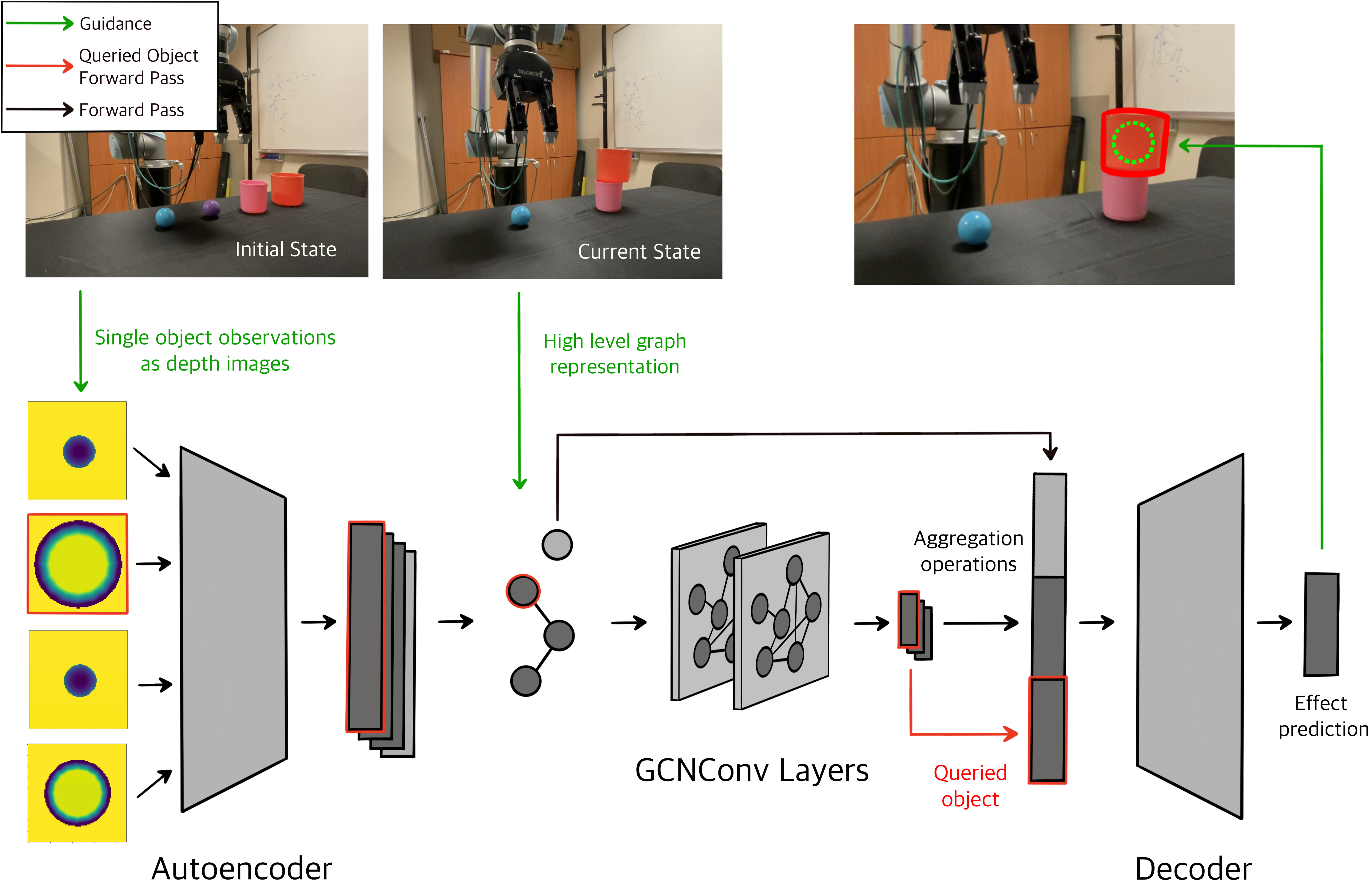}
    \caption{MOGAN: Multi-Object Graph Affordance Network Architecture, \tuba{along with the pretrained autoencoder}. The depth images of single objects are encoded with the autoencoder. It then constructs the graph representation of the compound object. The proposed model, MOGAN, extracts meaningful features from the graph and predicts the resulting effect between a single object and a queried object within the compound object. \reviewtwo{The predicted effect is visually depicted in the rightmost image with a dashed green circle.}}
    \label{fig:network}
\end{figure*}

\subsection{Graph Neural Networks}


\tuba{Graph Neural Networks have been used in a wide range of domains to model systems composed of multiple parts such as the human body for human action understanding \cite{shahid2022view}, the human hand for gesture inference \cite{li2021two}, the electroencephalogram (EEG)-related measurements for emotion recognition \cite{kong2022causal}, and the images with multiple entities for inferring the relations between them \cite{zhu2021multiscale}.} In robotics, on the other hand, the representation capability of GNNs for an unlimited number of objects and their relationships has enabled their widespread adoption \cite{iriondo2021affordance,tan2019object,zhu2021hierarchical,kulshrestha2023scl}. Lou et al. \cite{lou2022learning} depicted densely clustered objects as graph structures and extracted the adjacency and occlusion relations. They then utilized GNNs to learn grasping poses for target objects, taking into account the spatial relations with other objects. Wilson and Hermans  \cite{wilson2020learning} utilized GNNs in conjunction with CNNs to encode their multi-object environment for more accurate reward calculation during policy training. Lin et al. \cite{lin2022efficient} devised a graph structure in which objects and the goal positions for pick and place tasks are connected by edges. Subsequently, the learned GNN policy selects object and goal nodes to execute the steps of desired tasks.  Huang et al. \cite{huang2023planning}  represented multi-object scenes as fully connected graph structures based on partial observations and learned the relations between nodes as logical symbols using GNN classifiers. \tuba{In contrast, our study reasons relations between objects by learning observed effects of robot actions without defining logical symbols.}


GNNs are also commonly used for modeling dynamics of multi-object systems \cite{tian2023multi}. Driess et al. \cite{driess2023learning} employed GNNs to capture the dynamics between multiple objects for novel scene synthesis using Neural Radiance Fields (NeRF) \cite{mildenhall2021nerf}. Tekden et al. \cite{tekden2020belief,tekden2024object} introduced a learnable physics engine where objects are represented as graph structures, and the relations between them are classified at the edge level. With estimated relations between objects, future states are predicted based on the applied actions. However, the objects are simple-shaped, and their features are restricted to position and radius values. Additionally, only actions that push the objects in the horizontal plane are considered.


Overall, in our study, compound objects are represented as graph structures, their features are learned utilizing GNNs, and the affordances are learned through effect predictions. Our system plans a sequence of actions (selecting an object to place it on the compound object) with a search algorithm utilizing the learned affordances.

\section{Method}

Our proposed method  models the affordances of compound objects, which are composed of an arbitrary number of objects that are placed on top of each other. Given the compound object and a new object, it learns to predict the effects generated by placing the new object on top of the compound object. 
In our framework, an affordance, which is denoted as $A$, is defined as the relation between the compound object ($T$) that resides on the table, the object ($o$) to be placed on top of the compound object, and the effects ($E_1, E_2, E_3$) generated: $A = (T, o ,(E_1, E_2, E_3))$. Given $T$ and $o$, our system is expected to learn $E_1, E_2$, and $E_3$. For learning, at the start of each exploration cycle, the size of the object compound ($T$) is initialized as 0. Then, the robot randomly selects and picks up an object ($o$), places it on top of the current object compound, and observes the effects ($E_1, E_2, E_3$) until either the new object falls down or the object compound collapses.  

In the rest of this section, we first describe how compound and single objects ($T$ and $o$) and effects ($E_1, E_2, E_3$) are represented, and the details of the learning algorithm. Finally we describe how the learned affordances can be used to make plans in order to achieve different goals.

\subsection{Single Object Representation}

The single objects are represented by features extracted from their depth images using  an autoencoder. The encoder component of the autoencoder takes in a 32x32 normalized depth image and comprises three linear layers with neuron sizes of 256, 256, and 64, respectively. Empirically, a latent space size of 4 was found to be sufficient for representing the images of the object set used in this study. \tuba{The decoder part of the autoencoder is not utilized in the MOGAN model because we only need the latent vector; therefore, it is not shown in Figure \ref{fig:network}. The decoder depicted in the figure represents the decoder component of the MOGAN model.} The autoencoder is trained with the single depth images collected from both the simulation environment and the real world until convergence. 

\tuba{In our system, the encoder component of the autoencoder is used to extract latent representations from the single images. These latent representations are then used to construct graph representations for the compounds, allowing the MOGAN model to learn the effects from them.} The maximum and minimum values of the depth images are appended to the latent representations to prevent the information loss caused by the normalization operation.\tuba{Therefore, single objects $o$ and their features, such as size, shape, and concaveness, are represented by a learned feature vector of size 6.} 


\subsection{Compound Object Representation}
\label{CompoundObjectRepresentation}

The compound objects are composed of different objects placed on top of each other.  In order to represent a compound object, both the features of the single objects inside the compound and the spatial relations between the objects are required to be used. For this, we utilize a graph-based structure. A graph, denoted as $G$, is defined as a tuple of nodes $N$ and edges $E$.

\begin{equation}G = (N,E)\end{equation}
\begin{equation} N = \{ n_1,n_2, ..., n_k \}, n_i  \in \mathbb{R}^p, 1 \le i \le k 
\end{equation}
\begin{equation}
E = \{ e_1,e_2, ..., e_{k-1} \},  e_i  \in \mathbb{R}^q, 1 \le i \le k-1 . \end{equation}

Each node, $n$, consists of the object features acquired through the autoencoder. $p$ and $q$ denote the size of the feature vector of a node and an edge, respectively, while $k$ indicates the number of nodes within the graph, with no specific limits on this count. \reviewtwo{A directed edge between two nodes is defined when objects are placed consecutively in the tower, with the direction going from one object to the one placed before it}, providing a hint for the spatial relations between the objects in the compound. In Section \ref{building_nonlinear}, we also designed an edge creation algorithm for non-linear compounds to preserve these spatial relations. All nodes form self-connections.


\begin{figure}[h]
    \centering
    \includegraphics[ width=0.45\textwidth]{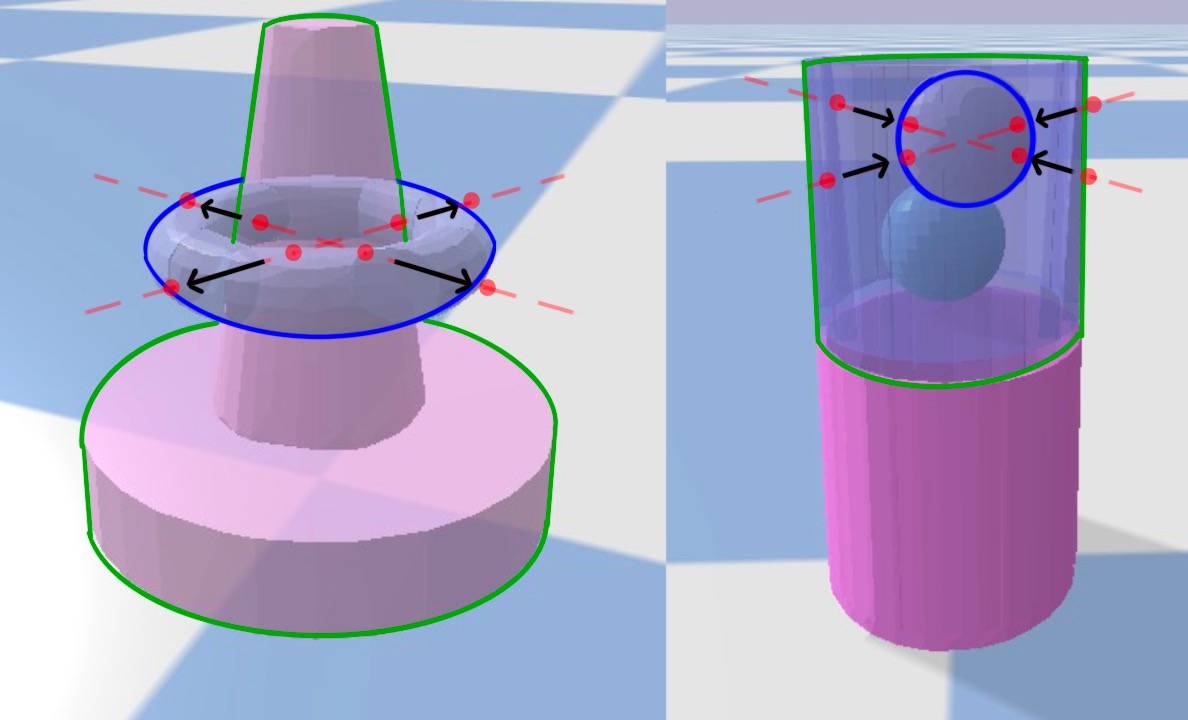}
    \caption{Visualization of the calculation of lateral spatial displacements: Imaginary rays are projected through the center of the new object. Red points illustrate the intersections with both the compounding object and newly added object. \reviewtwo{The black arrows are calculated by the function $s$.} }
    \label{fig:effect2}
\end{figure}

\subsection{Effect Representation}
When an object is placed on the compound object, different types of effects, such as insertion in different ways, stacking, or toppling, are observed. Instead of categorizing each effect instance into a pre-defined effect category, we propose a generic continuous effect representation that captures the 3D spatial relations between the placed object and each object in the compound.
In other words, the effect represents the spatial outcome of placing the new object on the compound and is encoded as a combination: $(E_1, E_2, E_3)$. $E_1$ describes the height differences between the top and bottom surfaces of each object pairs\tuba{, considering their bounding boxes.}
\begin{equation}
E_1 = \{ E_1^1, E_1^2, .., E_1^k \}, E_1^i \in \mathbb{R}^2, 1 \le i \le k \end{equation}
\begin{equation} E_1^i = \{s(|z_i^+-z_{k+1}^+|), s(|z_i^- -z_{k+1}^-|)\} \end{equation}
$E_1^i$ correpsponds to the effect between the new object and the $i^{\textrm{th}}$ object in the object compound. $z^+$ and $z^- $ describe the maximum and minimum height values of an object \reviewtwo{derived from the bounding box of the object}. $s$ is a sign function that assigns signs to the effect values. \reviewtwo{The function $s$ creates a vector starting from the top surface of the $i^{\textrm{th}}$ object to the top surface of the new object, using the center points of the surfaces. If the vector points toward the center of the $i^{\textrm{th}}$ object, the sign is considered negative. This comparison is performed for all the bounding box faces.} $E_2$ encodes the lateral spatial differences between objects. The differences are calculated by sending imaginary rays through the new object, as shown in Fig. \ref{fig:effect2}. If the ray does not intersect with the interested object (outlined with green color), the relevant effect becomes 0. The signs of the differences are calculated with the sign function $s$ considering the \tuba{points} that the imaginary rays cut. \tuba{The intersection points for each object are calculated using the $rayTestBatch$ function in PyBullet. This function provides the coordinates and object IDs where the ray intersects at each point. We use this function to find the red points indicated in Figure \ref{fig:effect2}.}
\begin{equation}E_2 = \{ E_2^1, E_2^2, .., E_2^k \}, E_2^i \in \mathbb{R}^4, 1 \le i \le k \end{equation}
\begin{equation} E_2^i = \{ s(|x_i^+-x_{k+1}^+|), s(|x_i^- -x_{k+1}^-|), \end{equation} 
\begin{equation} s(|y_i^+-y_{k+1}^+|),  s(|y_i^- -y_{k+1}^-|) \} \end{equation}

\begin{equation}
    E_3= 
\begin{cases}
    1,& \text{if } pos(o_i) \geq t_1 || ori(o_i) \geq t_2, 1 \le i \le k+1 \\
    0,              & \text{otherwise}
\end{cases}
\end{equation}

Finally, $E_3$ encodes whether the newly placed object falls down or the compound object collapses/topples when the new object is placed on top. The $pos$ and $ori$ functions get the \reviewtwo{x-y} position and orientation of a given object $o_i$ and compare them with the base of the compound. They return the sum of the differences in these values. The thresholds $t_1$ and $t_2$ are used to determine whether the position and orientation values indicate a collapse.\reviewtwo{They are chosen empirically as 20 cm and 60 degrees, respectively.}

\subsection{Multi-Object Graph Affordance Network (MOGAN)}
The proposed MOGAN model, shown in Fig. \ref{fig:network}, outputs the effects ($E_1, E_2, E_3$) expected to be generated when a new object ($o_{k+1}$) is placed on the compound object ($T$). As the compound object was formed by placing the objects one by one on top of each other, the depth images and the corresponding autoencoder features ($n_1, n_2, .. n_k$) were already collected and available for processing for a compound with a size of k. The autoencoder features of the new object to be placed on the compound object are also processed and is represented as $n_{k+1}$. Our system, MOGAN, comprises the encoder part of the pretrained autoencoder, GCNConv layers, and a linear decoder. The depth image features are extracted by the encoder. A high-level graph representation is then constructed based on the current compound ($T$), as explained in Section \ref{CompoundObjectRepresentation}, utilizing corresponding depth image features. Subsequently, two GCNConv \cite{kipf2016semi} layers process the graph representation of the compound to generate a latent representation. The mean and maximum values of these latent representations for all nodes are calculated and aggregated. The features of the newly added object ($n_{k+1}$), the aggregated latent representation, and the latent representation of the queried object are concatenated. The decoder, consisting of three linear layers, takes this concatenated input to predict the effects between the queried object ($o_{i}, 1 \le i \le k $) and the new object ($o_{k+1}$) placed on top of the compound. In our system, the parameter size for the network is 46786, and Leaky ReLU is utilized as the activation function between the layers.

\subsection{Planning and Tasks}

We aim to provide a variety of tasks to demonstrate the prediction capacity of the MOGAN for  planning to achieve different goals. The goals include obtaining object compounds of specified  heights, structures, and sub-structures. A tree search algorithm is realized to discover the optimal plan to achieve a specific goal. At each iterative step, the graph representation of the existing object compound is generated, and the object that will be placed on the tower is encoded. \tuba{Three MOGAN networks predict }the effects $E$ based on the graph representation of the compound and the feature vector of the new object. If predicted $E_3$ indicates a fall/collapse, the current branch of the search operation is terminated.

\begin{figure}[h]
    \centering
    \includegraphics[width=0.30\textwidth]{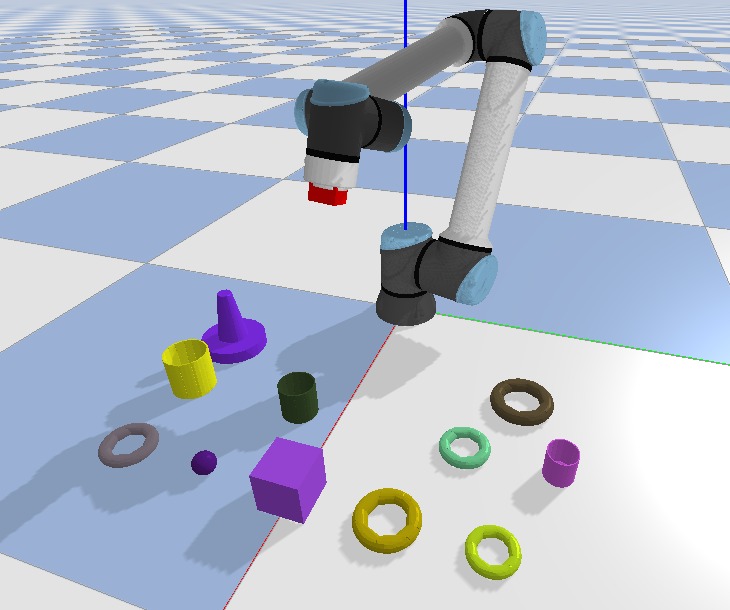}
    \caption{A PyBullet environment featuring a UR10 robot and various objects, including cubes, poles, balls, cups, and rings.}
    \label{fig:simenv}
\end{figure}

In detail, six different tasks can be specified. The first two tasks correspond to building the tallest and shortest compounds/towers. In order to predict the height of the object compounds, the $E_1$ effect predictions are summed up. The third and fourth tasks correspond to obtaining structures where the placed objects are required to enclose the top part of the object compound and become obstructed in the compound (inserted inside). The accumulated $E_2$ predictions are used for this purpose. The fifth task corresponds to building a tower of a specific height. Finally, the sixth task enables the selection of two objects from the set of objects that will be used in the object compound and puts constraints on their relative placements, such as maximizing or minimizing their relative distances.

\begin{figure}[h]
    \centering
    \includegraphics[ width=0.35\textwidth]{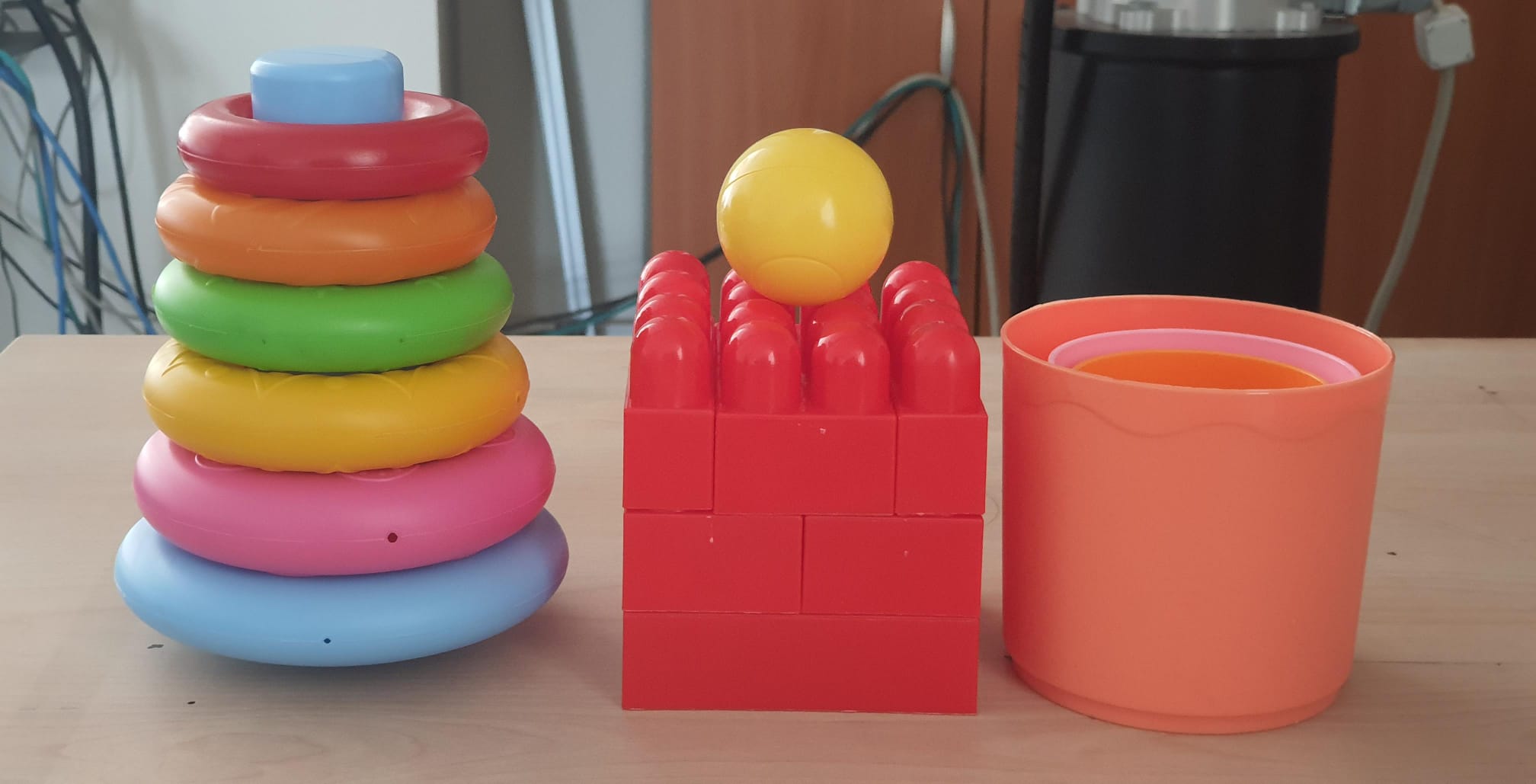}
    \caption{Various objects used in the real-world setup: a pole, rings, cups, a cube, and balls.}
    \label{fig:real_obj}
\end{figure}

\begin{table*}[t]
    \begin{center}
    \centering
    \caption{comparison of plan success rates in the simulation environment with the multi-object deepsym as baseline (mds)}
    \label{tab:plan_success}
    
\begin{tabular}{|l|ll|ll|ll|ll|ll|ll|}
\hline
\multirow{2}{*}{\textbf{Size}} & \multicolumn{2}{l|}{\textbf{Task 1: Tallest}} & \multicolumn{2}{l|}{\textbf{Task 2: Shortest}} & \multicolumn{2}{l|}{\textbf{Task 3: Occluded}} & \multicolumn{2}{l|}{\textbf{Task 4: Occluding}} & \multicolumn{2}{l|}{\textbf{Task 5: Specific Height}} & \multicolumn{2}{l|}{\textbf{Task 6: Condition}} \\ \cline{2-13} 
                                     & \multicolumn{1}{l|}{MOGAN}     & MDS     & \multicolumn{1}{l|}{MOGAN}      & MDS     & \multicolumn{1}{l|}{MOGAN}      & MDS     & \multicolumn{1}{l|}{MOGAN}      & MDS      & \multicolumn{1}{l|}{MOGAN}         & MDS         & \multicolumn{1}{l|}{MOGAN}      & MDS      \\ \hline
2                                    & \multicolumn{1}{l|}{100}        & 80            & \multicolumn{1}{l|}{100}         & 80            & \multicolumn{1}{l|}{100}          & 80           & \multicolumn{1}{l|}{100}          & 80             & \multicolumn{1}{l|}{100}             & 90               & \multicolumn{1}{l|}{100}          & 100             \\ \hline
3                                    & \multicolumn{1}{l|}{100}        & 60            & \multicolumn{1}{l|}{100}         & 80            & \multicolumn{1}{l|}{100}          & 90            & \multicolumn{1}{l|}{100}          & 70             & \multicolumn{1}{l|}{90}             & 60                & \multicolumn{1}{l|}{100}          & 60             \\ \hline
4                                    & \multicolumn{1}{l|}{90}         & 50            & \multicolumn{1}{l|}{90}          & 80            & \multicolumn{1}{l|}{80}          & 80            & \multicolumn{1}{l|}{90}          & 60             & \multicolumn{1}{l|}{90}             & 60                & \multicolumn{1}{l|}{90}          & 60             \\ \hline
5                                    & \multicolumn{1}{l|}{80}         & 10            & \multicolumn{1}{l|}{90}          & 60            & \multicolumn{1}{l|}{90}          & 40            & \multicolumn{1}{l|}{80}          & 60             & \multicolumn{1}{l|}{80}             & 50                & \multicolumn{1}{l|}{90}          & 60             \\ \hline

\end{tabular}
\end{center}
\end{table*}

\section{Experimental Setup}
\label{experimental_setup}

In the real-world experiment setup, we employ a 7-DOF UR10 manipulator equipped with a Robotiq 3-Finger Adaptive Robot Gripper. The objects chosen for the experiment are selected from a variety of toys commonly played with by infants, enabling the exploration of affordance relations involving concave, convex objects, and compounds. As indicated in Table \ref{tab:objects}, there are more small objects than large ones. This is because it becomes infeasible to find a suitable plan for the end-effector of the manipulator when dealing with compounds that are too tall. However, our approach does not impose a limitation on the number of objects in a compound.  Positioned 1 meter above the table, centered, a Realsense RGBD camera is installed, with its lens directed downward to optimize capture. The Pybullet environment is used for simulating actions and interactions. A custom gripper is attached to the wrist of the UR10 manipulator in the simulator in order to speed up the pick and place action executions. The objects used in the simulator are created using Blender, taking into account their size information. These objects are depicted in Fig. \ref{fig:real_obj} and Fig. \ref{fig:simenv}. The depth images of the simulated scene are captured using a virtual depth camera positioned at the same relative location as in the real-world experimental setup.

\begin{table}[]

\begin{center}
\centering
    \caption{object details}
    \label{tab:objects}

\begin{tabular}{|l|l|l|}

\hline
Name   & Size (Height, Width, Depth) (cm) & Number \\ \hline
Pole   & (17, 14, 14)                     & 1      \\ \hline
Ball & (5, 5, 5)                        & 5      \\ \hline
Cube   & (10, 10, 10)                     & 1      \\ \hline
Ring  & (3, 12, 12)                      & 1      \\ \hline
Ring  & (2.5, 10.5, 10.5)                & 1      \\ \hline
Ring  & (2.4, 9.7, 9.7)                  & 1      \\ \hline
Ring  & (2, 9, 9)                        & 1      \\ \hline
Ring  & (1.5, 8, 8)                      & 1      \\ \hline
Cup    & (10, 10.5, 10.5)                 & 1      \\ \hline
Cup    & (8.5, 7.5, 7.5)                  & 1      \\ \hline
Cup    & (7.5, 6.5, 6.5)                  & 1      \\ \hline
\end{tabular}
\end{center}
\end{table}

During experiments, a subset of the inventory is spawned in a rectangular area at random positions.   The depth image of the scene is segmented to acquire the depth images of the objects individually. In order to segment the depth image, the lowest values are grouped according to the pixel positions. The image is cropped according to the center pixel positions for each group. The positions of the objects are calculated using the center pixel positions and values and used during pick and place action executions. The trajectory for the manipulator to pick up and place an object is planned using MoveIt in the real-world setup, while built-in inverse kinematics functions are utilized in the simulation. The calculated positions for the objects are used as the positional goal for the end-effector during the pick operation. For the place operation, an additional 15 cm height is added to the goal position to eliminate potential object collisions. The orientation of the end-effector remains the same during operations.

A data point consists of: 1) a 32x32 single object depth image, 2) 32x32 individual depth images for the objects in the compound, and 3) effects as explained in the Method Section. The depth images for the objects in the compound are derived from previous iterations.  The training dataset comprises of 5000 data points acquired from simulation experiments. Prior to training, representations of both single and compound objects are acquired using the pretrained autoencoder.

 A MOGAN model is initialized with two GCNConv layers and three linear layers. The parameter size of the model is 46786 which is empirically found to prevent over fit. The model weights are randomly initialized with a torch seed value of 42. Mean Squared Error (MSE) loss and a custom sign loss are utilized as the loss functions. The sign loss, used for $E_1$ and $E_2$, penalizes predictions that do not align with the correct signs compared to the ground truth data. The Adam optimizer is employed as optimization algorithm. The model is trained for 600 epoch with a batch size of 1. The learning rate starts from $10^{-4}$ and gradually decreased with the learning rate scheduler. The gamma value is set to 0.95, and the step size is 500.

To demonstrate the efficiency of our proposed model, we compared it with a modified version of DeepSym \cite{ahmetoglu2022deepsym}. While Deepsym, a deep encoder-decoder network, learns relational symbols through effect predictions, the model is limited to paired object interactions. In the modified version, we encoded individual depth images and concatenated them. The size of the tensor is the multiplication of the feature size and the maximum object number in a compound. The maximum object number extracted from the dataset is 14 in our case. The remaining part of the input tensor remains 0 for the smaller-sized object compounds. In contrast to DeepSym, we did not utilize the Gumbel-Sigmoid function in the latent space, as our study does not focus on discrete symbol learning. Since we define our actions as adding a new object, we did not query additional actions. The decoder part learns the concatenation of all effects for each node in a compound.

 The parameter size for the baseline model is 50178, which is close to but not less than the parameter size of our proposed model. Training and test results are compared with the MOGAN model in the Experiments and Results Section.

\begin{figure}[h]
    \centering
    \includegraphics[width=0.45\textwidth]{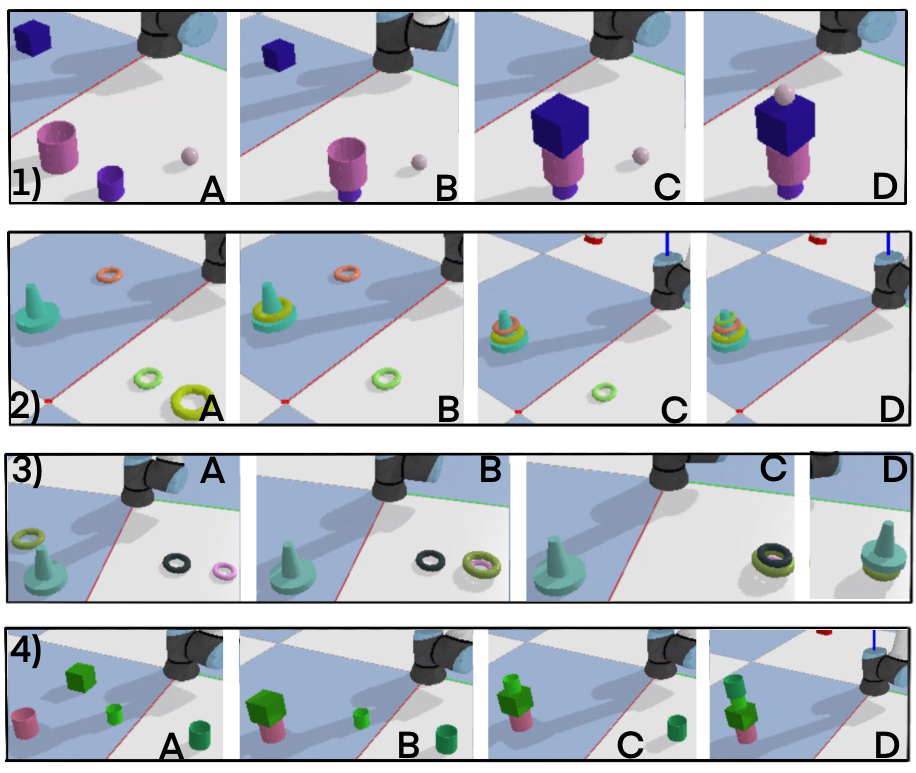}
    \caption{  A number of sample plan executions in the simulator.  \tba{ The tasks are (1) to minimize the invisibility of the given objects, (2) to build the shortest compound object using a pole and different sized rings, (3) to build the tallest compound object using a pole and different sized rings, and (4) to build a compound object given a constraint between the pink and dark green cups. } }
    \label{fig:5432sim}
\end{figure}

\begin{table}[t]
\begin{center}
    \centering
    \caption{prediction errors for the unseen simulation data \reviewtwo{in decimeters}}
    \label{tab:test_error}
\begin{tabular}{|l|lll|}
\hline
\multirow{2}{*}{Tower Size} & \multicolumn{3}{l|}{Test Errors }                                                                                                  \\ \cline{2-4} 
                            & \multicolumn{1}{l|}{Effect 1 (dm)} & \multicolumn{1}{l|}{Effect 2 (dm)} & Effect 3  \\ \hline
1                           & \multicolumn{1}{l|}{0.000}                        & \multicolumn{1}{l|}{0.000}                             & 0.010                \\ \hline
2                           & \multicolumn{1}{l|}{0.008}                        & \multicolumn{1}{l|}{0.000}                             & 0.099                \\ \hline
3                           & \multicolumn{1}{l|}{0.022}                        & \multicolumn{1}{l|}{0.004}                             & 0.149                \\ \hline
4                           & \multicolumn{1}{l|}{0.043}                        & \multicolumn{1}{l|}{0.003}                             & 0.177                \\ \hline
5                           & \multicolumn{1}{l|}{0.063}                        & \multicolumn{1}{l|}{0.002}                             & 0.177                \\ \hline
6                           & \multicolumn{1}{l|}{0.085}                        & \multicolumn{1}{l|}{0.002}                             & 0.184                \\ \hline
7                           & \multicolumn{1}{l|}{0.093}                        & \multicolumn{1}{l|}{0.001}                             & 0.240                \\ \hline
8                           & \multicolumn{1}{l|}{0.109}                        & \multicolumn{1}{l|}{0.001}                             & 0.145                \\ \hline
9                           & \multicolumn{1}{l|}{0.126}                        & \multicolumn{1}{l|}{0.000}                             & 0.185                \\ \hline
10                          & \multicolumn{1}{l|}{0.141}                        & \multicolumn{1}{l|}{0.000}                             & 0.231                \\ \hline
11                          & \multicolumn{1}{l|}{0.134}                        & \multicolumn{1}{l|}{0.000}                             & 0.306                \\ \hline
12                          & \multicolumn{1}{l|}{0.123}                        & \multicolumn{1}{l|}{0.000}                             & 0.322                \\ \hline
13                          & \multicolumn{1}{l|}{0.092}                        & \multicolumn{1}{l|}{0.000}                             & 0.330                \\ \hline
14                          & \multicolumn{1}{l|}{0.108}                        & \multicolumn{1}{l|}{0.000}                             & 0.252                \\ \hline
\end{tabular}
\end{center}
\end{table}

\section{Results}

\begin{figure*}[t]
    \centering
    \includegraphics[width=17cm]{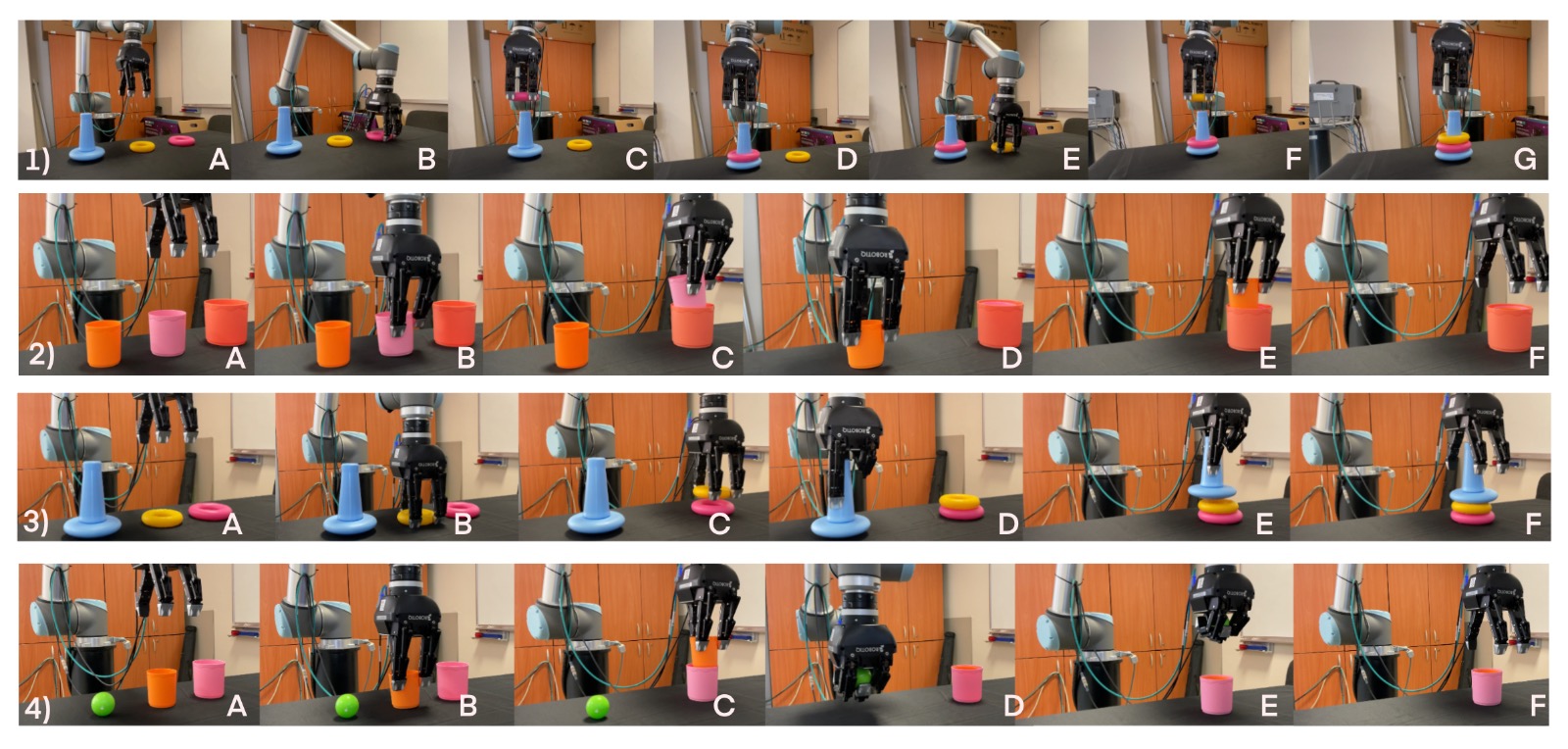}
    \caption{A number of snapshots from real-world planning experiments. In the first, second, and fourth images, the objective is to construct the shortest compound objects. In the third image, the goal is to create the tallest compound object. The system observes the scene, predicts the effects of each potential plan using MOGAN, and executes the optimal one.}
    \label{fig:real_world_execution}
\end{figure*}

In this section, we analyze the prediction error of our model for the unseen combinations of the composite objects and provide the results in Table \ref{tab:test_error}. The errors are grouped according to the compound object sizes to analyze the relation between compound object sizes and prediction errors. Effect 1 is the predicted height differences between two objects, as explained in the Method Section. The inventory contains objects with maximum, minimum, and mean height values of 17 cm, 1.5 cm, and 6.5 cm, respectively. The errors in Effect 1 predictions result in deviations of less than 1 cm in predicted height differences when the compound object size is 8 or less. If the object size exceeds 8, we observe a maximum error of 1.41 cm. Although these prediction errors do not significantly impact the majority of predictions due to the presence of considerably larger objects, they can lead to failures when predicting effects between smaller objects, such as small rings. The error for Effect 2 does not increase along with the compound object size. We can confidently state that our model is capable of predicting x and y displacements of objects without being affected by the number of objects. The ground truth value of Effect 3 is 1 when the tower collides, 0 otherwise. When we inspect the prediction errors for Effect 3, we see that it increases as the number of objects increases. However, the errors in Effect have minimal impact on the overall results due to the margin between ground truth values.

\subsection{ Simulation Experiments \& Comparison with Baseline}
We evaluate the generated plans for six different tasks, we sample 10 different configurations for each compound object size, ranging from 2 to 5 in the simulator\reviewtwo{, as shown in Table \ref{tab:plan_success}}. For the fifth task, which is to build a compound object with a desired height, we calculated possible height values for the sampled configuration, selected one as the goal, and compared it to the resulting height. For the last task, we randomly selected two objects from the sampled set of objects to maximize or minimize their distances. Please see the generated and executed plans for a number of sample tasks in Fig \ref{fig:5432sim}. \tba{In the 2nd and 3rd rows of Fig \ref{fig:5432sim}, different tasks are assigned for the same set of objects. In the 2nd row, the model benefits the \textit{passability} of rings onto the pole to keep the compound short. In the 3rd row, the model first benefits the \textit{stackability} of rings and then stacks the pole to increase the height of the compound.}

\begin{figure}[h]
    \centering
    \includegraphics[width=0.45\textwidth]{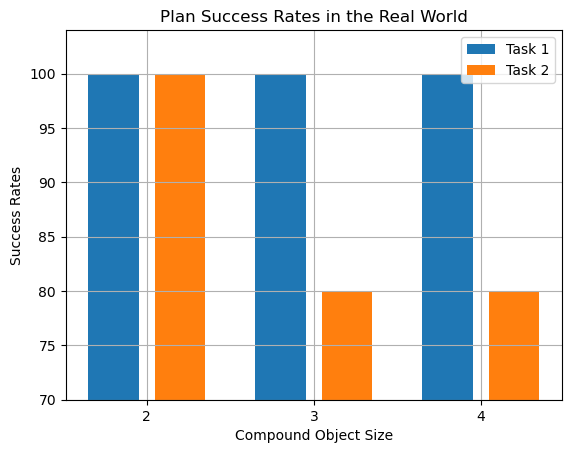}
    \caption{Plan success rates in the real world. The goals are to build the shortest and tallest compound objects. 5 trials were conducted for each set of different sizes.}
    \label{fig:rwes}
\end{figure}

Out of 300 planning tasks, our system was able to generate 283 successful plans, as shown in Table \ref{tab:plan_success}. The success rate was observed to slightly drop when the number of objects increases. This was an expected result, as the number of objects in the compound increase, predicting the affordance of the compound object and how it is affected from placing another object on top becomes more difficult. Additionally, as the number of objects increases, the number of predictions done during the planning increases exponentially. One erroneous prediction among all the correct predictions may cause a failure in planning. It is important to note that our MOGAN model significantly outperformed the base model in planning, as shown in Table \ref{tab:plan_success}, showing the effectiveness of using graph structures where the features of the objects in the compound are embedded in the nodes of the GNN for modeling multi-object affordances and for the multi-object planning problems.

\begin{figure}[h]
    \centering
    \includegraphics[width=0.45\textwidth]{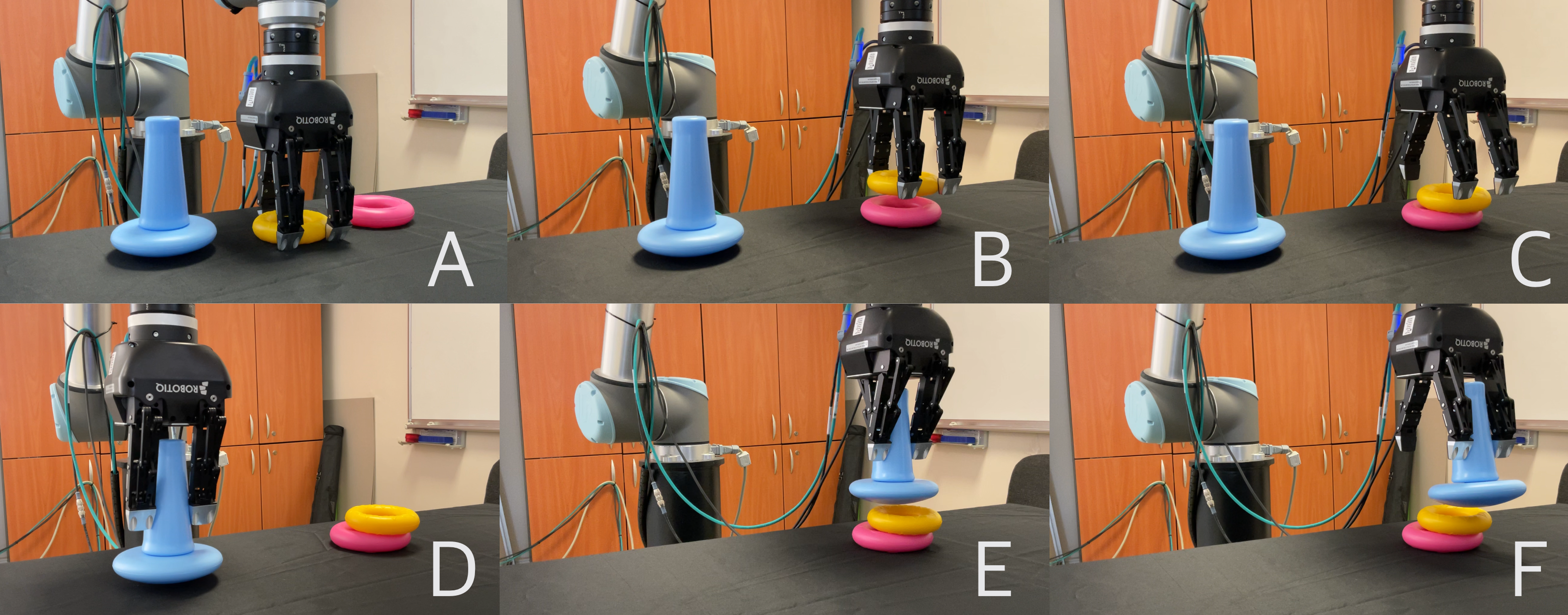}
    \caption{\reviewtwo{A failure case is illustrated here. When the goal is to build the tallest compound, the MOGAN model chooses to place the yellow ring and the pole on top of the pink ring, respectively. However, the pole gets squeezed between the fingers of the 3-finger gripper, resulting in failure, as shown in part F.}}
    \label{fig:real_failure}
\end{figure}

\subsection{Real-world Experiments}
In the real-world setup, we test our system's planning capacity with the first two tasks: building the shortest and the tallest compound objects. We sampled 5 sets of objects for the compound object sizes 2, 3, and 4. A number of plan execution snapshots from sampled tasks are provided in Fig.\ref{fig:real_world_execution}. Out of the 30 real-world planning tasks, 28 of the generated plans were found to be successful, as shown in Figure \ref{fig:rwes}. The system is able to build desired compound objects 1) using the depth images from Realsense, 2) predicting effects with the MOGAN model, 3) planning an optimal path with the tree search algorithm, and 4) executing it with the UR10 manipulator. The success rate slightly decreases as the object number in the inventory increases. Another reason for the failure is the unpredictability of the real-world systems. \reviewtwo{Since the objects we use are plastic, they exhibit slight elasticity. Therefore,  the object models in the simulation do not fully capture the physical properties of real-world objects. This can lead to unexpected results during the gripper's open and close operations. In Figure \ref{fig:real_failure}, }when the robot holds the pole, it grips it too tightly. As a result, when the gripper opens, the pole gets stuck between the fingers and does not fall.
\subsection{Building Nonlinear Compounds}
\label{building_nonlinear}

\tuba{In this section, we analyze experiments involving additional stacking actions. These actions include changes in position and orientation. In this setup, the agent can choose from 3 different locations along the x-axis and 2 different orientations (upside and downside), resulting in a total of 6 different actions. Orientation information is included in the feature vector to encode the objects' orientation, thus generating node features. To connect the nodes, edges are created between them following the Algorithm \ref{alg:alg_example}. $checkIntersections$ function returns $True$ if the bounding boxes of the objects intersect. The $checkConnectivity$ function determines whether two objects are in contact. If either of these functions returns $True$, an edge is added between the two objects.}

\begin{algorithm}
\caption{Edge Creation}\label{alg:alg_example}
\begin{algorithmic}[1]
\State List of object IDs $id\_list$
\State $edge\_list \gets [\ ]$
\For{$i = 1$ to $length(id\_list) - 1$}
    \State $a \gets id\_list[i]$
    \State $b \gets id\_list[i+1]$
    \State $intersections \gets checkIntersections(a, b)$
    \State $connectivity \gets checkConnectivity(a, b)$
    \If{$intersections$ or $connectivity$}
        \State append($edge\_list$, $(a, b)$)
    \EndIf
\EndFor
\end{algorithmic}
\end{algorithm}

\begin{table*}[]
\begin{center}

\caption{prediction errors for the unseen simulation data \reviewtwo{in decimeters} }
    \label{tab:ablation_test_error}

\begin{tabular}{|l|lllllllll|}
\hline
\multirow{3}{*}{Tower Size} & \multicolumn{9}{l|}{Test Errors}                                                                                                                                                                                                                                               \\ \cline{2-10} 
                            & \multicolumn{3}{l|}{Effect 1 (dm)}                                                                      & \multicolumn{3}{l|}{Effect 2 (dm)}                                                     & \multicolumn{3}{l|}{Effect 3}                                               \\ \cline{2-10} 
                            & \multicolumn{1}{l|}{MOGAN}          & \multicolumn{1}{l|}{MOGAAN}         & \multicolumn{1}{l|}{MOFFAN} & \multicolumn{1}{l|}{MOGAN} & \multicolumn{1}{l|}{MOGAAN} & \multicolumn{1}{l|}{MOFFAN} & \multicolumn{1}{l|}{MOGAN}           & \multicolumn{1}{l|}{MOGAAN} & MOFFAN \\ \hline
1                           & \multicolumn{1}{l|}{0.036}          & \multicolumn{1}{l|}{0.037}          & \multicolumn{1}{l|}{0.035}  & \multicolumn{1}{l|}{0.000} & \multicolumn{1}{l|}{0.000}  & \multicolumn{1}{l|}{0.000}  & \multicolumn{1}{l|}{0.013}           & \multicolumn{1}{l|}{0.010}  & 0.011  \\ \hline
2                           & \multicolumn{1}{l|}{\textbf{0.041}} & \multicolumn{1}{l|}{0.042}          & \multicolumn{1}{l|}{0.050}  & \multicolumn{1}{l|}{0.001} & \multicolumn{1}{l|}{0.002}  & \multicolumn{1}{l|}{0.001}  & \multicolumn{1}{l|}{0.019}           & \multicolumn{1}{l|}{0.0535} & 0.015  \\ \hline
3                           & \multicolumn{1}{l|}{\textbf{0.023}} & \multicolumn{1}{l|}{\textbf{0.023}} & \multicolumn{1}{l|}{0.156}  & \multicolumn{1}{l|}{0.000} & \multicolumn{1}{l|}{0.000}  & \multicolumn{1}{l|}{0.001}  & \multicolumn{1}{l|}{0.0521} & \multicolumn{1}{l|}{0.204}  & 0.033  \\ \hline
4                           & \multicolumn{1}{l|}{\textbf{0.044}} & \multicolumn{1}{l|}{0.073}          & \multicolumn{1}{l|}{0.067}  & \multicolumn{1}{l|}{0.000} & \multicolumn{1}{l|}{0.001}  & \multicolumn{1}{l|}{0.000}  & \multicolumn{1}{l|}{\textbf{0.186}}  & \multicolumn{1}{l|}{0.253}  & 0.323  \\ \hline
\end{tabular}
                      
\end{center}
\end{table*}

\begin{table}[]
\begin{center}
 \caption{comparison of plan success rates with mogaan and moffan}
    \label{tab:ablation_plan_success}
    
\begin{tabular}{|l|l|l|l|}
\hline
       & Planning Success & No Solution & Failure \\ \hline
MOGAN  & 100              & 0           & 0       \\ \hline
MOGAAN & 60               & 40          & 0       \\ \hline
MOFFAN & 70               & 0           & 30      \\ \hline
\end{tabular}
\end{center}
\end{table}

\tuba{In this setup, two additional models are designed for comparison with the proposed MOGAN model. The first is the Multi-Object Graph Attentive Affordance Network (MOGAAN), where we utilize GATConv layers instead of GCNConv layers. The second is the Multi-Object Feed Forward Network (MOFFAN), where we use linear layers for encoding. In the latter, we concatenate the object features according to the object selection order. To train and test the models with the new actions, we collected a dataset of 3000 compounds consisting of cubes and cups, using 500 of them for testing. The MOGAN model is initialized as described in Section \ref{experimental_setup}, with the input sizes adjusted to match the feature sizes of the dataset, and the x-axis position conditioned in the latent space. The parameter sizes for the other models are 47618, 48290 respectively. The training parameters for this experiment are also described in Section  \ref{experimental_setup}.  }

\begin{figure}[h]
    \centering
    \includegraphics[width=0.38\textwidth]{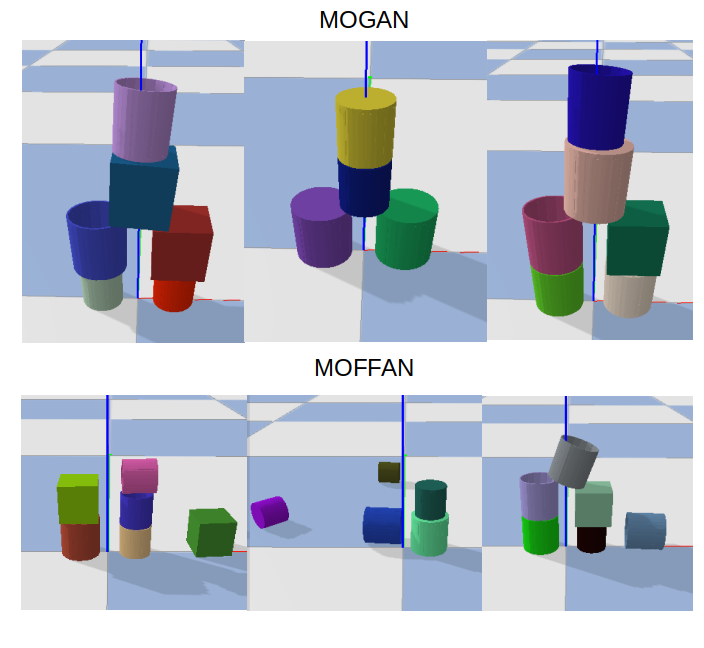}
    \caption{Three experiments where the MOFFAN model fails are shown in comparison to the MOGAN model. The MOGAN model can reason about the lengths of the legs of a compound to successfully stack another object onto the middle, whereas the MOFFAN model cannot.}
    \label{fig:moffan_failure}
\end{figure}

The test errors for this experiment are provided in Table \ref{tab:ablation_test_error}. The errors are grouped according to the compound object sizes. The errors for predicted height differences are shown under the Effect 1 column, where the Effect 2 column corresponds to the errors for lateral differences, and the Effect 3 column corresponds to the prediction errors for collapse. The comparisons with MOGAAN and MOFFAN indicate that our model outperforms the baselines, especially as the compound size increases.

\tuba{For this experiment, we designed a task where the goal is to build a compound shaped like a bridge. When a list of 6 objects is presented, the models are required to find the stacking order and orientations for the predefined locations that result in a bridge shape. Out of 10 planning tasks, our proposed model, MOGAN, was able to plan compounds for all the desired tasks, and the plans were executed successfully in the simulation environment. The MOGAAN model could not generate plans for 4 of the tasks. The MOFFAN model generated 10 plans, but 3 of them collapsed in the simulation environment as shown in Table \ref{tab:ablation_plan_success}. Figure \ref{fig:moffan_failure} shows a comparison of the plans where the MOFFAN fails. While the MOGAN model builds legs of the bridge with similar sizes to stack another object in the middle, the MOFFAN model cannot build legs with similar sizes. This demonstrates that the MOFFAN model lacks the capability to reason about the relationship between leg sizes and stability, whereas our model can reason about the affordances of the bridge legs to construct a stable bridge.}

\tuba{In the following part of this section, we analyze two case studies to better understand the capabilities of the MOGAN model.}

\begin{figure*}[t]
\begin{center}

    \centering
    \includegraphics[width=13cm]{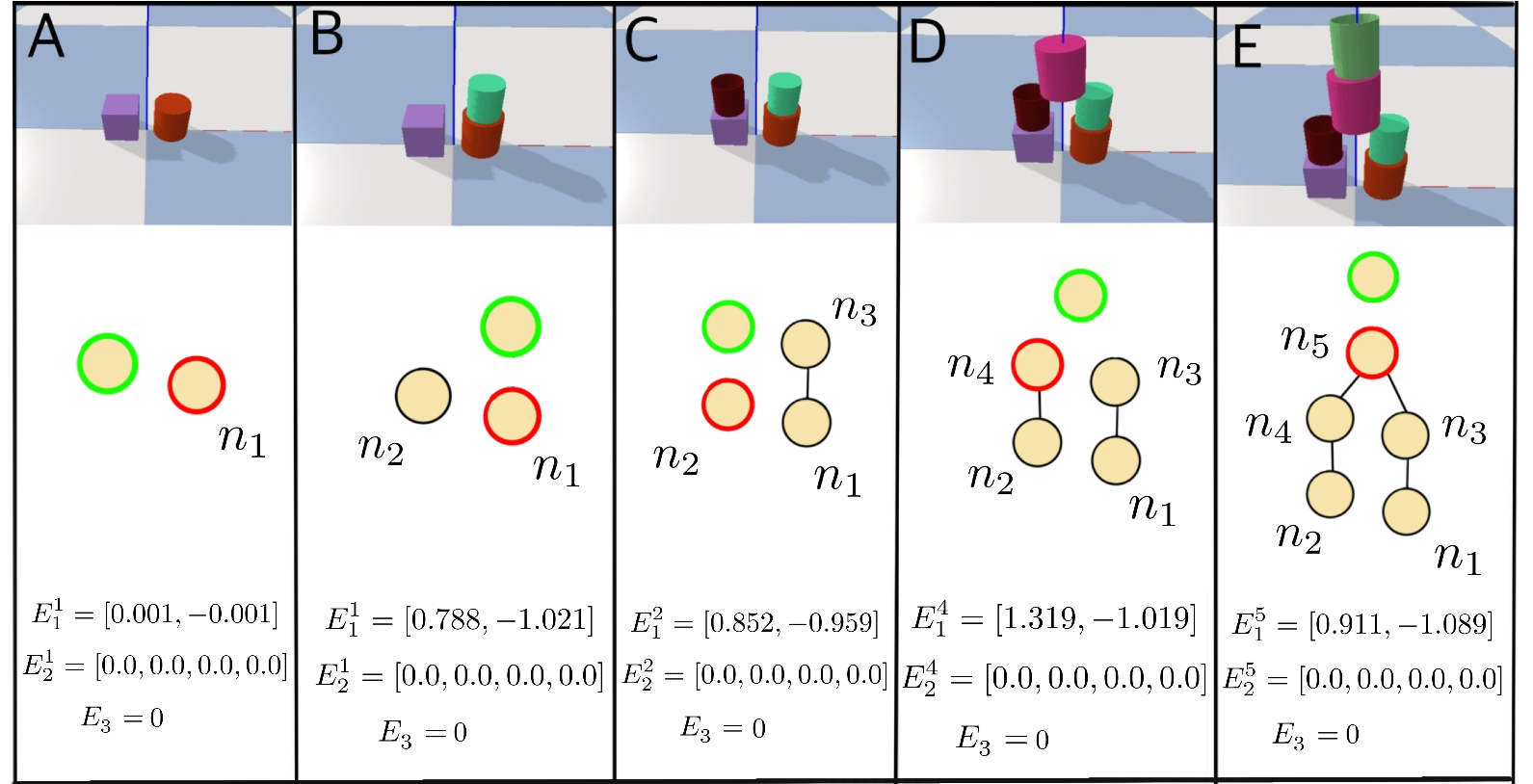}
    \caption{An example of online graph generation and effect prediction for a new test set containing six different stacking actions is shown. The rotation of the objects is encoded in the node features, while the x-axis position is conditioned in the latent space. In the image, the effects of the green-circled object corresponding to the red-circled objects are displayed. }
    \label{fig:bridge}
    \end{center}
\end{figure*}

\subsubsection{Case Study 1}

\tuba{ A compound building process is shown in Figure \ref{fig:bridge}, along with depiction of the graph structures and predicted effects. The figure shows the effect predictions for the new objects (green circles) in relation to the objects corresponding to the red-circled nodes. In part B of the figure, a cup is stacked onto a rotated cup. The model can reason about the rotations of objects, leading to accurate effect predictions. In parts C,D, and E, it is shown that the MOGAN model can reason about the spatial relations of objects in graphs with various edge connections, leading to accurate effect predictions when a new object is to be stacked.}

\subsubsection{Case Study 2}

\begin{figure*}[]
    \centering
    \includegraphics[width=11cm]{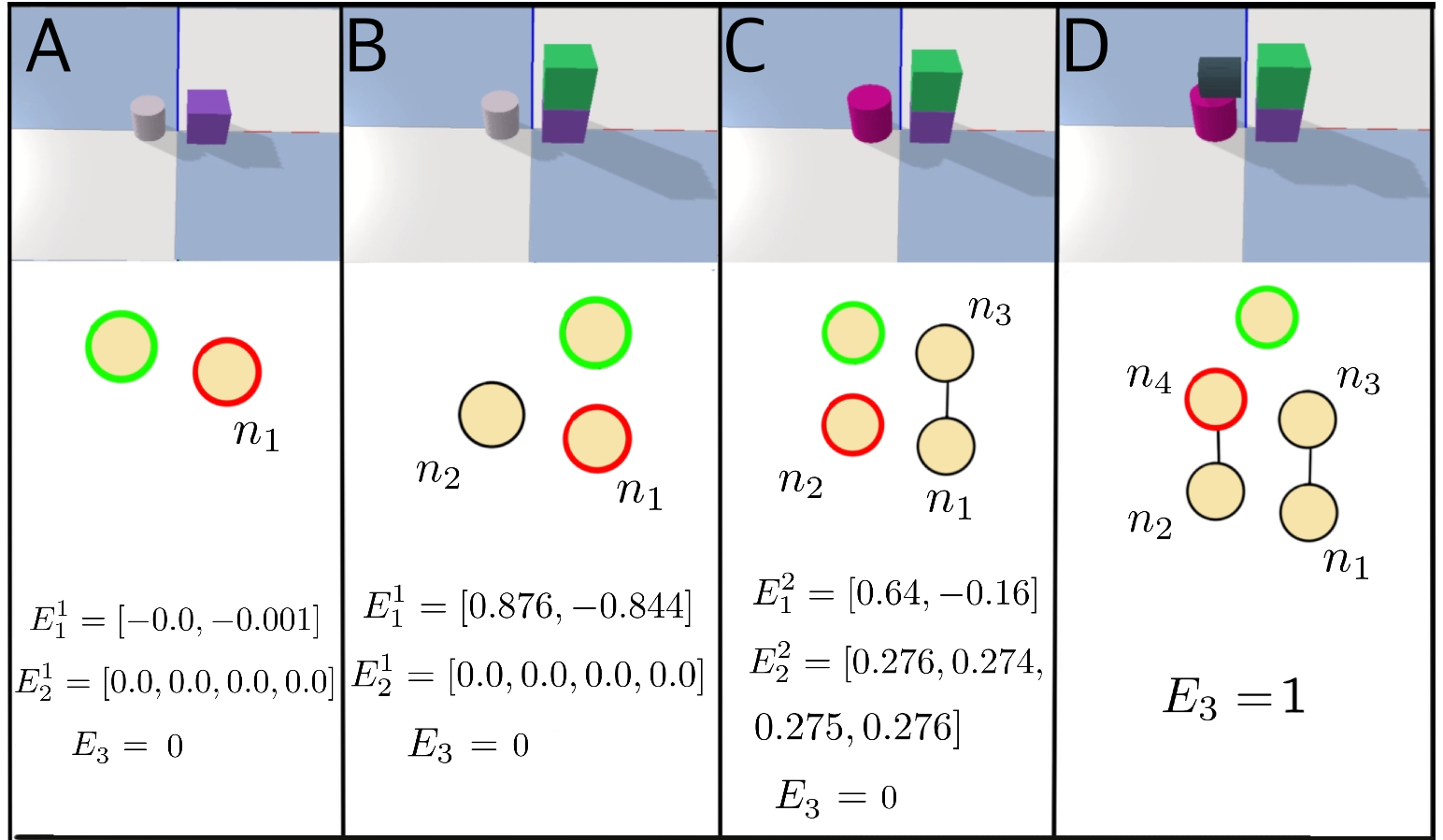}
    \caption{An example of online graph generation and effect prediction for collision detection is provided. Since the pink cup covers the gray cup, the lengths of the compound's legs vary, preventing further stacking.}
    \label{fig:covering}
\end{figure*}

\tuba{In this experiment, while building a compound, a rotated cup covers a smaller cup as shown in part C of Figure  \ref{fig:covering}. Note that for all objects with different orientations, depth images are taken in the initial position and features are extracted by the autoencoder. Then, the object's orientation information is appended to the feature vector. The MOGAN model is able to learn the continuous effects $E_1$ and $E_2$ for the ``covering" effect, as shown in part C of the figure, where the predicted effects indicate intersecting bounding boxes, with the top and side of the red-circled object remaining inside the newly added green-circled object. Additionally, the model predicts effect $E_3$ as $True$ before the stacking action when the height difference between the subcompounds is too large to place a new object in the middle, demonstrating its ability to reason about spatial relations between nodes in a graph to predict the collapse effect, as shown in part D of Figure \ref{fig:covering}.}

\section{Conclusion}

In this research, we  proposed a novel Multi-Object Graph Affordance Network, MOGAN, which models affordances of compound objects for manipulation and planning. 
We showed that our system was able to correctly predict the affordances of compound objects that include spheres, cups, poles, and several rings that enclose the poles. This prediction capability was effectively used to build different structures via planning structures of highly complex affordances. 
\tba{In the future, we plan to discover symbolic affordances of compound structures and utilize AI planners for task realization. Additionally, to enable an end-to-end and generalizable approach, we intend to use point clouds \tuba{and RGB images} to represent our objects and depth values to retrieve effect of actions.}

\bibliographystyle{IEEEtran}
\bibliography{references}



\end{document}